\documentclass{svproc}
\usepackage{url}

\usepackage{graphicx}
\usepackage{makecell}
\usepackage{sidecap}
\usepackage{amsmath}
\usepackage{multirow}
\usepackage{amssymb}
\usepackage{hyperref}
\usepackage{algorithm}
\usepackage{algorithmicx}
\usepackage{algpseudocode}
\usepackage{booktabs}
\usepackage{wrapfig}
\usepackage{enumitem}

\begin{document}
\mainmatter              
\title{DAN: Decentralized Attention-based Neural Network for the MinMax Multiple Traveling Salesman Problem}
\titlerunning{DAN: Decentralized Attention-based Neural Network for the MinMax mTSP}  
%
\author{Yuhong Cao\inst{1} \and Zhanhong Sun\inst{1} \and Guillaume Sartoretti\inst{1}}
%
\authorrunning{Yuhong Cao, Zhanhong Sun, and Guillaume Sartoretti}
%
\tocauthor{Yuhong Cao, Zhanhong Sun, and Guillaume Sartoretti}
\institute{National University of Singapore, Mechanical Engineering Dept., Singapore,\\
\email{caoyuhong@u.nus.edu, sun\_z@u.nus.edu, guillaume.sartoretti@nus.edu.sg}}

\maketitle              

\begin{abstract}
The multiple traveling salesman problem (mTSP) is a well-known NP-hard problem with numerous real-world applications. In particular, this work addresses MinMax mTSP, where the objective is to minimize the max tour length among all agents. Many robotic deployments require recomputing potentially large mTSP instances frequently, making the natural trade-off between computing time and solution quality of great importance. However, exact and heuristic algorithms become inefficient as the number of cities increases, due to their computational complexity. Encouraged by the recent developments in deep reinforcement learning (dRL), this work approaches the mTSP as a cooperative task and introduces DAN, a decentralized attention-based neural method that aims at tackling this key trade-off. In DAN, agents learn fully decentralized policies to collaboratively construct a tour, by predicting each other's future decisions. Our model relies on the Transformer architecture and is trained using multi-agent RL with parameter sharing, providing natural scalability to the numbers of agents and cities. Our experimental results on small- to large-scale mTSP instances ($50$ to $1000$ cities and $5$ to $20$ agents) show that DAN is able to match or outperform state-of-the-art solvers while keeping planning times low. In particular, given the same computation time budget, DAN outperforms all conventional and dRL-based baselines on larger-scale instances (more than 100 cities, more than 5 agents), and exhibits enhanced agent collaboration. A video explaining our approach and presenting our results is available at \url{https://youtu.be/xi3cLsDsLvs}.
\end{abstract}
\section{INTRODUCTION}
\label{DARS2022-DAN—INTRODUCTION}

The traveling salesman problem (TSP) is a challenging NP-hard problem, where given a group of cities (i.e., nodes) of a given complete graph an agent needs to find a complete \textit{tour} of this graph, i.e., a closed path from a given starting node that visits all other nodes exactly once with minimal path length. The TSP can be further extended to the multiple traveling salesman problem (mTSP), where multiple agents collaborate with each other to visit all cities from a common starting node with minimal cost. In particular, MinMax (minimizing the max tour length among agents, i.e., total task duration) and MinSum (minimizing total tour length) are two most common objectives for mTSP~\cite{kaempfer2018learning,hu2020reinforcement,park_schedulenet_2021}. Although simple in nature, the TSP and mTSP are ubiquitous in robotics problems that need to address agent distribution, task allocation, and/or path planning, such as multi-robot patrolling, last mile delivery, or distributed search/coverage. For example, Faigl et al.~\cite{faigl2012goal}, Obwald et al.~\cite{osswald2016speeding} and Cao et al.~\cite{cao2021tare} proposed methods for robot/multi-robot exploration tasks, at the core of which is such a TSP/mTSP instance, whose solution quality will influence the overall performance. More generally, dynamic environments are involved in many robotic deployments, where the underlying graph may change with time and thus require the TSP/mTSP solution be recomputed frequently and rapidly (within seconds or tens of seconds if possible). While state-of-the-art exact algorithms (e.g., CPLEX~\cite{CPLEX}, LKH3~\cite{helsgaun2017extension}, and Gurobi~\cite{Gurobi}) can find near-optimal solutions to TSP instances in a few seconds, exact algorithms become unusable for mTSP instances if computing time is limited~\cite{hu2020reinforcement}. As an alternative, meta-heuristic algorithms like \textit{OR Tools}~\cite{ORtools} were proposed to balance solution quality and computing time. In recent years, neural-based methods have been developed to solve TSP instances~\cite{vinyals2015pointer,bello2016neural,kool2018attention} and showed promising advantages over heuristic algorithms both in terms of computing time and solution quality. However, neural-based methods for mTSP are scarce\cite{park_schedulenet_2021,hu2020reinforcement}. In this work, we introduce DAN, a decentralized attention-based neural approach for the MinMax mTSP, which is able to quickly and distributedly generate tours that minimize the time needed for the whole team to visit all cities and return to the depot (i.e., the \textit{makespan}).

\begin{wrapfigure}{r}{0.3\textwidth}
    \vspace{-0.6cm}
    \centering
    \includegraphics[width=0.3\textwidth]{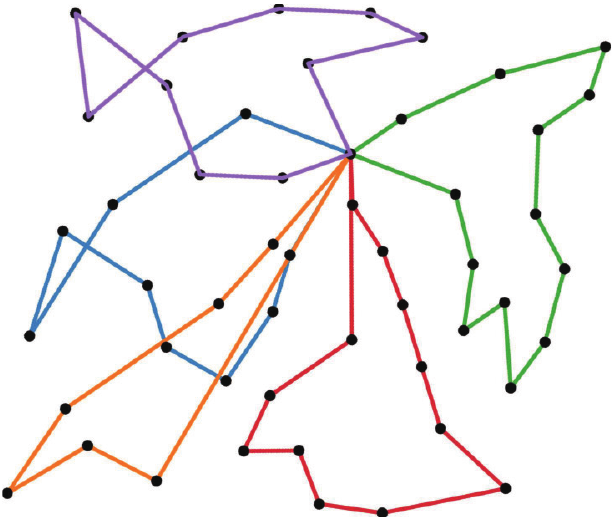}
    \vspace{-0.5cm}
    \caption{DAN's final solution to an example mTSP problem.}
    \label{figure1}
    \vspace{-0.65cm}
\end{wrapfigure}
We focus on solving mTSP as a decentralized cooperation problem, where agents each construct their own tour towards a common objective. To this end, we rely on a threefold approach: first, we formulate mTSP as a sequential decision-making problem where agents make decisions asynchronously towards enhanced collaboration. Second, we propose an attention based neural network to allow agents to make individual decisions according to their own observations, which provides agents with the ability to implicitly predict other agents' future decisions, by modeling the dependencies of all the agents and cities. Third, we train our model using multi-agent reinforcement learning with parameter sharing, which provides our model with natural scalability to arbitrary team sizes.

We present test results on randomized mTSP instances involving 50 to 1000 cities and 5 to 20 agents, and compare DAN's performance with that of exact, meta-heuristic, and dRL methods. Our results highlight that DAN achieves performance close to \textit{OR Tools}, a highly optimized meta-heuristic baseline~\cite{ORtools}, in relatively small-scale mTSP (fewer than 100 cities). In relatively large-scale mTSP, our model is able to significantly outperform \textit{OR Tools} both in terms of solution quality and computing time. We believe this advantage makes DAN more reliable in dynamic robotic tasks than non-learning approaches. DAN also outperforms two recent dRL based methods in terms of solution quality for nearly all instances.

\section{PRIOR WORKS}
\label{DARS2022-DAN-priorWorks}


For TSP, exact algorithms like dynamic programming and integer programming can theoretically guarantee optimal solutions. However, these algorithms do not scale well.
Nevertheless, exact algorithms with handcrafted heuristics (e.g., \textit{CPLEX}~\cite{CPLEX}) remain state-of-the-art, since they can reduce the search space efficiently.
Neural network methods for TSP became competitive after the recent advancements in dRL.
Vinyals et al.~\cite{vinyals2015pointer} first built the connection between deep learning and TSP by proposing the Pointer network, a sequence-to-sequence model with a modified attention mechanism, which allows one neural network to solve TSP instances composed of an arbitrary number of cities. 
Kool et al.~\cite{kool2018attention} replaced the recurrent unit in the Pointer Network and proposed a model based on a Transformer unit~\cite{vaswani2017attention}.
Taking the advantage of self-attention on modeling the dependencies among cities, Kool et al.'s achieved significant improvement in term of solution quality. 


In mTSP, cities need to be partitioned and allocated to each agent in addition to finding optimal sequences/tours. This added complexity makes state-of-the-art exact algorithm TSP solvers impractical for larger mTSP instances. 
\textit{OR Tools}~\cite{ORtools}, developed by Google, is a highly optimized meta-heuristic algorithm. Although it does not guarantee optimal solutions, OR Tools is one of the most popular mTSP solvers because it effectively balances the trade-off between solution quality and computing time. 
A few recent neural-based methods have also approached the mTSP in a decentralized manner. Notably, Hu et al.~\cite{hu2020reinforcement} proposed a model based on a shared graph neural network and distributed attention mechanism networks to first allocate cities to agents. \textit{OR Tools} can then be used to quickly generate the tour associated with each agent.
Park et al.~\cite{park_schedulenet_2021} presented an end-to-end decentralized model based on graph attention network, which could solve mTSP instances with arbitrary numbers of agents and cities. They used a type-ware graph attention mechanism 
to learn the dependencies between cities and agents, where the extracted agents' feature and cities' features are then concatenated during the embedding procedure before outputting the final policy.


\section{PROBLEM FORMULATION}
\label{DARS2022-DAN-problemFormulation}

The mTSP is defined on a graph $G=(V,E)$, where $V=\left\{1,...,n\right\}$ is a set of $n$ nodes (cities), and $E$ is a set of edges. In this work, we consider $G$ to be \textit{complete}, i.e., $(i,j)\in E$ for all $i \neq j$. Node $1$ is defined as the depot, where all $m$ agents are initially placed, and the remaining nodes as cities to be visited. The cities must be visited exactly once by any agent. After all cities are visited, all agents return to the depot to finish their tour.  Following the usual mTSP statement in robotics~\cite{bektas_multiple_2006}, we use the Euclidean distance between cities as edge weights, i.e., this work addresses the \textit{Euclidean mTSP}.
We define a solution to mTSP $\pi = \left\{\pi^1,...,\pi^m \right\}$ as a set of agent tours. Each agent tour $\pi^i=(\pi^i_1,\pi^i_2,...,\pi^i_{n_i})$ is an ordered set of the cities visited by this agent, where $\pi^i_t \in V$ and $\pi^i_1=\pi^i_{n_i}$ is the depot. $n_i$ denotes the number of cities in this agent tour, so $\sum_{i=1}^{m}n_i=n+2m-1$ since all agent tours involve the depot twice.
Denoting the Euclidean distance between cities $i$ and $j$ as $c_{ij}$, the cost of agent $i$'s tour reads: $L(\pi^i)=\sum \limits_{j=1}^{n_i-1}c_{\pi^i_j\pi^i_{j+1}}$.
As discussed in Section~\ref{DARS2022-DAN—INTRODUCTION},  we consider MinMax (minimize $\max \limits_{i\in\{1,...,m\}}L(\pi^i)$) as the objective of our model.


\section{mTSP AS A RL PROBLEM}
\label{DARS2022-DAN-mTSPAsARLProblem}

In this section, we cast mTSP into a decentralized multi-agent reinforcement learning (MARL) framework. In our work, we consider mTSP as a cooperative task instead of an optimization task. We first formulate mTSP as a sequential decision making problem, which allows RL to tackle mTSP. We then detail the agents' state and action spaces, and the reward structure used in our work.


\begin{description}[leftmargin=*]

\item[\textbf{Sequential Decision Making}]

Building upon recent works on neural-based TSP solvers, we consider mTSP as a sequential decision-making problem. That is, we let agents interact with the environment by performing a sequence of decisions, which in mTSP are to select the next city to visit. These decisions are made sequentially and asynchronously by each agent based on their own observation, upon arriving at the next city along their tour, thus constructing a global solution collaboratively and iteratively. Each decision (i.e., RL action) will transfer the agent from the current city to the next city it selects. 
Assuming all agents move at the same, uniform velocity, each transition takes time directly proportional to the Euclidean distance between the current city and the next city. We denote the remaining time until the next decision of agent $i$ as $g_i$ (i.e., its remaining travel time). In practice, we discretize time in steps, and let $g_i$ decrease by $\Delta g$ at each time step, which defines the agents' velocity. This assumption respects the actual time needed by agents to move about the domain, but also lets agents make decisions asynchronously (i.e., only when they reach the next city on their tour). We note that staggering the agents' decision in time naturally helps avoid potential conflicts in the city selection process, by allowing agents to select cities in a sequentially-conditional manner (i.e., agents select one after the other, each having access to the decisions already made by agents before them in this sequence).
We empirically found that this significantly improves collaboration and performance. \\[-0.3cm]

\begin{algorithm}[t]
    \caption{Sequential decision making to solve mTSP.}
    \label{alg1}
    \renewcommand{\algorithmicrequire}{\textbf{Input:}}
    \renewcommand{\algorithmicensure}{\textbf{Output:}}
        \begin{algorithmic}
                  \Require number of agents $m$, graph $G = (V, E)$
        \Ensure Solution $\pi$
        \State Initialize mask $M = \{0\}$, remaining travel time $g = \{0\}$,
        and \\empty tours starting at the depot $\pi^i = \{1\}$ ($1 \leq i \leq m$).
        \While{${\rm sum}(M) < n$}
        \For {$i=1,...,m$}
            \If {$g_i \leq 0$}
            \State  Observe $s_i^c$, $s_i^a$ of agent $i$ and outputs $p$
            \State Select next city $\pi^{i}_t$ from $p$ ($t={\rm Length}(\pi^i)+1$)
            \State Append $\pi^{i}_t$ to $\pi^i$,  $M[\pi^{i}_t] \gets 1$, $g_i \gets c_{\pi^{i}_{t-1} \pi^{i}_{t}}$
            
            \EndIf
        \State $g_i \gets g_i - \Delta g$
        \EndFor
        \EndWhile
        \State \textbf{return} $\pi = \{\pi^1,...,\pi^m\}$
        \end{algorithmic}
\end{algorithm}

\item[\textbf{Observation}]

We consider a fully observable world where each agent can access the states of all cities and all agents. Although a partial observation is more common in decentralized MARL~\cite{zhang_multi-agent_2021}, a global observation is necessary to make our model comparable to baseline algorithms, and partial observability will be considered in future works.
Each agent's observation consists of three parts: the cities state, the agents state, and a global~mask.

The cities state $s^{c}_{i}=(x^{c}_{i},y^{c}_{i}), i\in\{1,...,n\}$ contains the Euclidean coordinates of all cities relative to the observing agent. Compared to absolute information, we empirically found that relative coordinates can help prevent premature convergence and lead to a better final policy.

The agents state $s^{a}_{i}=(x^{a}_{i},y^{a}_{i},g_{i})$, $ i\in\{1,...,m\}$ contains the Euclidean coordinates of all agents relative to the observing agent, and the agents' remaining travel time $g_{i}$. As mTSP is a cooperative task, one agent can benefit from observing the state of other agents, e.g., to predict their future decisions.

Finally, agents can observe a global mask $M$: an $n$-dimensional binary vector containing the visit history of all $n$ cities. Each entry of $M$ is initially $0$, and is set to 1 after any agent has visited the corresponding city. Note that the depot is always unmasked during the task. This help agents avoid to be forced to visit remaining cities even it would lead to worse solutions. \\[-0.3cm]


\item[\textbf{Action}]

At each decision step of agent $i$, based on its current observation $(s^c_i, s^a_i, M)$, our decentralized attention-based neural network outputs a stochastic policy $p(\pi^i_t)$, parameterized by the set of weights $\theta$:
$p_\theta(\pi^i_t = j | s^c_i, s^a_i, M)$, where $j$ denotes an unvisited city. Agent $i$ takes an action based on this policy to select the next city $\pi^i_t$. By performing such actions iteratively, agent $i$ constructs its tour $\pi^i$. \\[-0.3cm]


\item[\textbf{Reward Structure}]

To show the advantage of reinforcement learning, we try to minimize the amount of domain knowledge introduced into our approach. In this work, the reward is simply the negative of the max tour length among agents: $R(\pi)=-\max \limits_{i\in\{1,..,m\}}(L(\pi^i))$, and all agents share it as a global reward. This reward structure is sparse, i.e., agents only get rewarded after all agents finish their tours.

\end{description}

\section{DAN: DECENTRALIZED ATTENTION-BASED NETWORK}
\label{DARS2022-DAN-DecentralizedAttentionBasedNeuralNetwork}

We propose an attention-based neural network, composed of a city encoder, an agent encoder, a city-agent encoder, and a decoder.
Its structure is used to model three kinds of dependencies in mTSP, i.e., the agent-agent dependencies, the city-city dependencies, and the agent-city dependencies.
To achieve good collaboration in mTSP, it is important for agents to learn all of these dependencies to make decisions that benefit the whole team.
Each agent uses its local DAN network to select the next city based on its own observation.
Compared to existing attention-based TSP solvers, which only learn dependencies among cities and finds good individual tours, DAN further endows agents with the ability to predict each others' future decision to improve agent-city allocation, by adding the agent and the city-agent encoders.

Fig.~\ref{figure2} shows the structure of DAN.
Based on the observations of the deciding agent, we first use the city encoder and the agent encoder to model the dependencies among cities and among agents respectively.
In the city-agent encoder, we then update the city features by considering other agents' potential decisions according to their features.
Finally, in the decoder, based on the deciding agent's current state and the updated city features, we allocate attention weights to each city, which we directly use as its policy.


\begin{description}[leftmargin=*]

\item[\textbf{Attention Layer}]

The Transformer attention layer~\cite{vaswani2017attention} is used as the fundamental building block in our model. The input of such an attention layer consists of the query source $h^q$ and the key-and-value source $h^{k,v}$, which are both vectors with the same dimension. The attention layer updates the query source using the weighted sum of the value, where the attention weight depends on the similarity between query and key. We compute the query $q_{i}$, key $k_{i}$ and value $v_{i}$ as: $q_{i}=W^Qh^{q}_{i}, k_{i}=W^kh^{k,v}_{i}, v_{i}=W^vh^{k,v}_{i}$, where $W^Q, W^K, W^V$ are all learnable matrices with size $d\times d$. Next, we compute the similarity $u_{ij}$ between the query $q_i$ and the key $k_j$ using a scaled dot product: $u_{ij}=\frac{q_{i}^T\cdot k_{j}}{\sqrt{d}}$.
Then we calculate the attention weights $a_{ij}$ using a softmax: $a_{ij}=\frac{e^{u_{ij}}}{\sum_{j=1}^{n}e^{u_{ij}}}$.
Finally, we compute a weighted sum of these values as the output embedding from this attention layer: $h_{i}^{'}=\sum_{j=1}^{n}a_{ij}v_{j}$.
The embedding content is then passed through the feed forward sublayer (containing two linear layer and a ReLU activation).
Layer normalization and residual connections are used within these two sublayers as in~\cite{vaswani2017attention}. \\[-0.3cm]

\begin{figure}[tb]
  \centering
  \includegraphics[width=\textwidth]{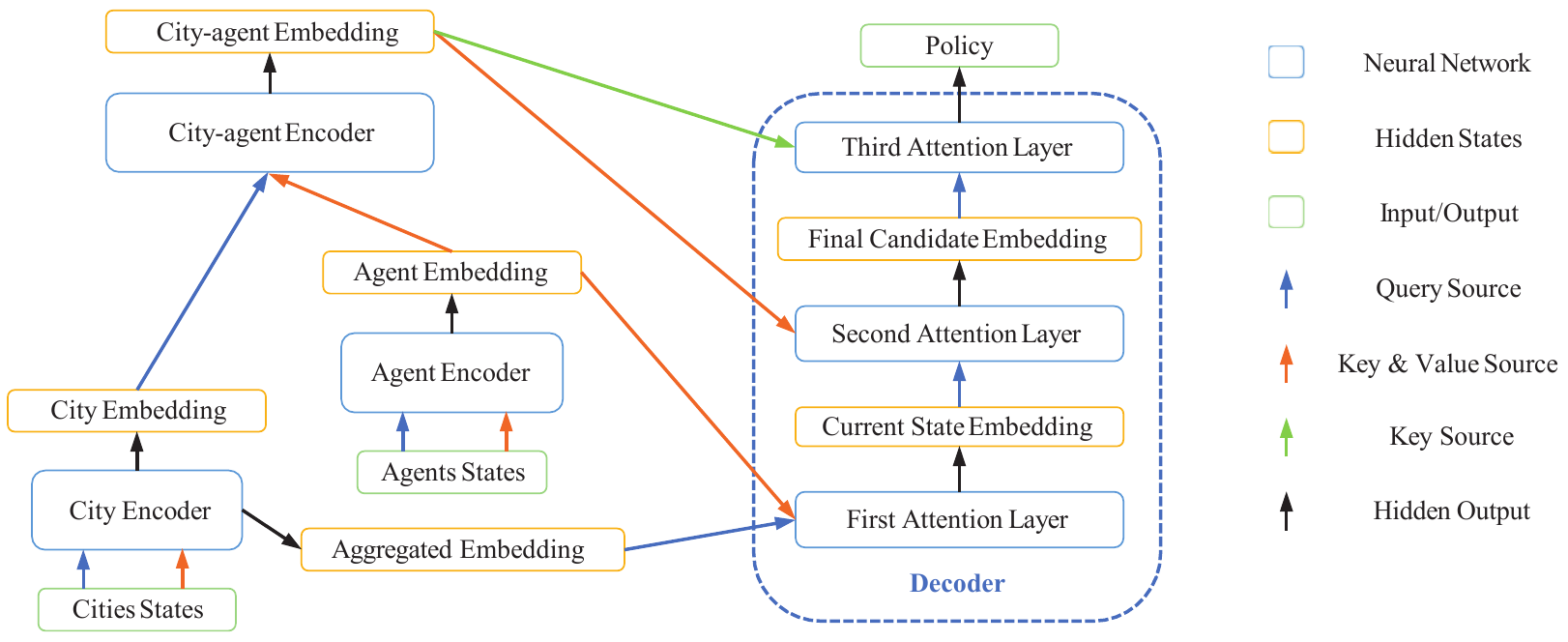}
  \caption{DAN consists of a city encoder, an agent encoder, a city-agent encoder and a final decoder, which allows each agent to individually process its inputs (the \textit{cities states}, and the \textit{agents states}), to finally obtain its own city selection policy. In particular, the agent and city-agent encoders are introduced in this work to endow agents with the ability to predict each others’ future decision and improve the decentralized distribution of agents.}
  \label{figure2}
\end{figure}


\item[\textbf{City Encoder}]

The city encoder is used to extract features from the cities state $s^c_i$ and model their dependencies. The city encoder first embeds the relative Euclidean coordinates $x^{c}_{i}$ of city $i$, $i\in\left\{2,...,n\right\}$ into a $d$-dimensional ($d=128$ in practice) initial city embedding $h^{c}_{i}$ using a linear layer. Similarly, the depot's Euclidean coordinates $x^{c}_{1}$ are embedded by another linear layer to $h^{c}_{1}$. The initial city embedding is then passed through an attention layer. Here $h^q=h^{k,v}=h^{c}$, as is commonly done in self-attention mechanisms. Self-attention achieved good performance to model the dependencies of cities in single TSP approaches~\cite{kool2018attention}, and we propose to rely on the same fundamental idea to model the dependencies in mTSP.
We term the output of the city encoder, $h^{'c}$, the \textit{city embedding}.
It contains the dependencies between each city $i$ and all other cities. \\[-0.3cm]


\item[\textbf{Agent Encoder}]

The agent encoder is used to extract features from the agents state $s^a_i$ and model their dependencies.
A linear layer is used to separately embed each (3-dimensional) component of $s^a_i$ into the initial agent embedding $h^{a}_{i}$.
This embedding is then passed through an attention layer, where $h^q=h^{k,v}=h^{a}$. We term the output of this encoder, $h^{'a}$ the \textit{agent embedding}. It contains the dependencies between each agent $i$ and all other agents. \\[-0.3cm]


\item[\textbf{City-agent Encoder}]

The city-agent encoder is used to model the dependencies between cities and agents. The city-agent encoder applies an attention layer with \textit{cross-attention}, where $h^q=h^{'c}, h^k{k,v}=h^{'a}$.
We term the output $h^{ca}$ of this encoder the \textit{city-agent embedding}. It contains the relationship between each city $i$ and each agent $j$ and implicitly predicts whether city $i$ is likely to be selected by another agent $j$, which is one of the keys to the improved performance of our model. \\[-0.3cm]


\item[\textbf{Decoder}]

The decoder is used to decode the different embeddings into a policy for selecting the next city to visit. The decoder starts with encoding the deciding agent's current state. We choose to express the current agent state implicitly by computing an aggregated embedding $h^s$ which is the mean of the city embedding. This operation is similar to the graph embedding used in~\cite{kool2018attention}.

The first attention layer then adds the agent embedding to the aggregated embedding. In doing so, it relates the state of the deciding agent to that of all other agents. Here $h^q=h^s$ and $ h^{k,v}=h^{'a}$. This layer outputs the current state embedding $h^{'s}$. 
After that, a second attention layer is used to compute the final candidate embedding $h^{f}$, where $h^q=h^{'s}, h^{k,v}=h^{ca}$. This layer serves as a \textit{glimpse} which is common to improve attention mechanisms~\cite{bello2016neural}.
There, when computing the similarity, we rely on the global mask $M$ to manually set the similarity $u_i=-\infty$ if the corresponding city $i$ has already been visited to ensure the attention weights of visited cities are $0$.
The final candidate embedding $h^{f}$ then passes through a third and final attention layer.
The query source is the final candidate embedding $h^{f}$, and the key source is the city-agent embedding $h^{ca}$.
For this final layer only only, following~\cite{vinyals2015pointer}, we directly use the vector of attention weights as the final policy for the deciding agent.
The same masking operation is also applied in this layer to satisfy the mTSP constraint. 
These similarities are normalized using a Softmax operation, to finally yield the probability distribution $p$ for the next city to visit: $p_i=p_\theta(\pi^j_t=i|s^c_i, s^a_i, M)={e^{u_i}}/{\sum_{i=1}^{n}e^{u_i}}$.

\end{description}


\section{TRAINING}
\label{DARS2022-DAN-Training}

In this section, we describe how DAN is trained, including the choice of hyperparameters and hardware used.


\begin{description}[leftmargin=*]

\item[\textbf{REINFORCE with Rollout Baseline}]

In order to train our model, we define the policy loss: $L=-\mathbf{E}_{p_{\theta}(\pi^i)}[R(\pi)]$, where $p_{\theta}(\pi^i)=\prod \limits_{t=1}^{n_i}p_\theta(\pi^i_t|s^c_i, s^a_i, M)$. The policy loss is the expectation of the negative of the max length among the tours of agents. 
The loss is optimized by gradient descent using the REINFORCE algorithm with a greedy rollout baseline~\cite{kool2018attention}.
That is, we re-run the same exact episode from the start a second time, and let all agents take decisions by \textit{greedily} exploiting the best policy so far (i.e., the ``baseline model'' explained in Section~\ref{DARS2022-distributed-training} below).
The cumulative reward $b(\pi)$ of this baseline episode is then used to estimate the advantage function: $A(\pi) = R(\pi) - b(\pi)$ (with $R(\pi)$ the cumulative reward at each state of the RL episode).
This helps reduce the gradient variance and avoids the burden of training the model to explicitly estimate the state value, as in traditional actor-critic algorithms.
The final gradient estimator for the policy loss reads:
$L = -\mathbf{E}_{p_{\theta}(\pi^i)}[(R(\pi)-b(\pi))\nabla {\rm log}p_\theta (\pi^i)]$. \\[-0.3cm]


\item[\textbf{Distributed Training}]
\label{DARS2022-distributed-training}

We train our model using parameter sharing, a general method for MARL~\cite{gupta_cooperative_nodate}. That is, we allow agents to share the parameters of a common neural network, thus making the training more efficient by relying on the sum of experience from all agents. Our model is trained on a workstation equipped with a i9-10980XE CPU and four NVIDIA GeForce RTX 3090 GPUs. We train our model utilizing Ray, a distributed framework for machine learning~\cite{moritz2018ray} to accelerate training by parallelizing the code. With Ray, we run 8 mTSP instances in parallel and pass gradients to be applied to the global network under the A2C structure~\cite{noauthor_openai_2017}.
At each training episode, the positions of cities are generated uniformly at random in the unit square $[0,1]^2$ and the velocity of agents is set to $\Delta g=0.1$. The number of agent is randomized within $[5,10]$ and the number of cites is randomized within $[20,100]$ during early training, which needs 96 hours to converge. After initial convergence of the policy, the number of cities is randomized within $[20,200]$ for further refinement, requiring 24 hours of training. We formulate one training batch after 8 mTSP instances are solved, and perform one gradient update for each agent. We train the model with the Adam optimizer~\cite{kingma_adam_2017} and use an initial learning rate of $10^{-5}$ and decay every 1024 steps by a factor of $0.96$.
Our full training and testing code is available at \url{https://bit.ly/DAN_mTSP}.

\end{description}


\section{Experiments}
\label{DARS2022-DAN-experiments}

We test our decentralized attention-based neural network (DAN) on numerous sets of 500 mTSP instances each, generated uniformly at random in the unit square $[0,1]^2$. 
We test two different variants of our model, which utilize the same trained policy differently:
\begin{itemize}
    \item Greedy: each agent always selects the action with highest activation in its policy.
    \item Sampling: each agent selects the city stochastically according to its policy.
\end{itemize}
For our sampling variant, we run our model multiple times on the same instance and report the solution with the highest quality. While~\cite{kool2018attention} sample 1280 solutions for single TSP, we only sample 64 solutions (denoted as s.64) for each mTSP instance to balance the trade-off between computing time and solution quality. 
In the test, the velocity of agents is set to $\Delta g=0.01$ to improves the performance of our model by allowing more finely-grained asynchronous action selection.


\subsection{Results}

\begin{table}[t]\footnotesize
    \centering
    \caption{Results on random mTSP set (500 instances each). $n$ denotes the number of cities
    and $m$ denotes the number of agents}
    \vspace{-0.3cm}
    \begin{tabular}{p{1.5cm}|p{0.7cm}p{0.7cm}|p{0.7cm}p{0.7cm}|p{0.7cm}p{0.7cm}|p{0.7cm}p{0.7cm}|p{0.7cm}p{0.7cm}}
    \toprule
         \multicolumn{1}{c|}{\multirow{2}{*}{Method}} & \multicolumn{2}{c|}{n=50 m=5} & \multicolumn{2}{c|}{n=50 m=7} & \multicolumn{2}{c|}{n=50 m=10} & \multicolumn{2}{c|}{n=100 m=5} & \multicolumn{2}{c}{n=100 m=10} \\
           & Max. & T(s) & Max. & T(s) & Max. & T(s) & Max. & T(s) & Max. & T(s)\\
         \midrule
         EA    & 2.35 & 7.82 & 2.08 & 9.58 & 1.96 & 11.50  & 3.55 & 12.80  & 2.75 & 17.52 \\
         SOM   & 2.57 & 0.76 & 2.30 & 0.78 & 2.16 & 0.76 & 3.10 & 1.58 & 2.41 & 1.58 \\
         OR & \textbf{2.04} & 12.00 & \textbf{1.96} & 12.00 & 1.96 & 12.00 & \textbf{2.36}  & 18.00 & 2.29 & 18.00 \\
         Gurobi & 2.54 & 3600 &  &  & 2.42 & 3600 & 7.29 & 3600 & 7.17 & 3600 \\
         \midrule
         Kool et al. & 2.84 & 0.20 & 2.64 & 0.22 & 2.52 & 0.25 & 3.28 & 0.37 & 2.77 & 0.41 \\
         Park et al. & 2.37 &       & 2.18 &       & 2.10 &  & 2.88 &       & 2.23 &  \\
         Hu et al. & 2.12 & 0.01 & & & 1.95 & 0.02 & 2.48 & 0.04 & 2.09 & 0.04 \\
         \midrule
         DAN(g.) & 2.29 & 0.25 & 2.11 & 0.26 & 2.03 & 0.30 & 2.72 & 0.43 & 2.17 & 0.48 \\
         DAN(s.64) & 2.12 & 7.87 & 1.99 & 9.38 & \textbf{1.95} & 11.26 & 2.55 & 12.18 & \textbf{2.05} & 14.81 \\
         \bottomrule
    \end{tabular}
    \begin{tabular}{p{1.5cm}|p{0.7cm}p{0.7cm}|p{0.7cm}p{0.7cm}|p{0.7cm}p{0.7cm}|p{0.7cm}p{0.7cm}}
         \toprule
         \multicolumn{1}{c|}{\multirow{2}{*}{Method}} 
         & \multicolumn{2}{c|}{n=100 m=15} &
          \multicolumn{2}{c|}{n=200 m=10} & \multicolumn{2}{c|}{n=200 m=15} & \multicolumn{2}{c}{n=200 m=20} \\
          & Max. & T(s) & Max. & T(s) & Max. & T(s) & Max. & T(s)\\
         \midrule
         EA   & 2.51 & 21.63 & 4.07 & 29.91 & 3.62 & 34.33 & 3.37 & 39.34\\
         SOM  & 2.22 & 1.57 & 2.81 & 3.01 & 2.50 & 3.04 & 2.34 & 3.04\\
         OR   & 2.25 & 18.00  & 2.57 & 63.70 & 2.59 & 60.29 & 2.59 & 61.74 \\
         \midrule
         Kool et al. & 2.64 & 0.46  & 3.27 & 0.78 & 2.92 & 0.83 & 2.77 & 0.89 \\
         Park et al. & 2.16 &  & 2.50 &       & 2.38 &       & 2.44 &  \\
         \midrule
         DAN(g.) & 2.09 & 0.58 & 2.40 & 0.93 & 2.20 & 0.98 & 2.15 & 1.07 \\
         DAN(s.64) & \textbf{2.00} & 19.13 & \textbf{2.29} & 23.49 & \textbf{2.13} & 26.27 & \textbf{2.07} & 29.83 \\
         \bottomrule
    \end{tabular}
    \label{table1}
\end{table}

\begin{table*}[t]
    \centering
    \caption{Results on the large-scale mTSP set (500 instances each) where the number of agents is fixed to 10. }
    \vspace{-0.3cm}
    \begin{tabular}{l|p{0.75cm}p{0.75cm}|p{0.75cm}p{0.75cm}|p{0.75cm}p{0.75cm}|p{0.75cm}p{0.75cm}|p{0.75cm}p{0.75cm}|p{0.75cm}p{0.75cm}}
    \toprule
         \multicolumn{1}{c|}{\multirow{2}{*}{Method}} & \multicolumn{2}{c|}{n=500} & \multicolumn{2}{c|}{n=600} & \multicolumn{2}{c|}{n=700} & \multicolumn{2}{c|}{n=800} & \multicolumn{2}{c|}{n=900} & \multicolumn{2}{c}{n=1000}\\
         & Max. & T(s) & Max. & T(s) & Max. & T(s) & Max. & T(s) & Max. & T(s)& Max. & T(s) \\
         \midrule
         OR \hfill\cite{hu2020reinforcement} & 7.75& 1800 & 9.64 & 1800 & 11.24 & 1800 & 12.34 & 1800 & 13.71 & 1800 & 14.84 & 1800 \\
         SOM \hfill\cite{lupoaie_som-guided_2019} & 3.86 & 7.63 & 4.24 & 9.39 & 4.54 & 10.86 & 4.93 & 14.28 & 5.21 & 16.65 & 5.53 & 17.89 \\
         Hu et al. \hfill\cite{hu2020reinforcement} & 3.32 & \textbf{0.56} & 3.65 & \textbf{0.81} & 3.95 & \textbf{1.22} & 4.20 & \textbf{1.69} & 4.59 & \textbf{2.21} & 4.81 & \textbf{2.87} \\
         DAN(g.) & 3.29 & 2.15 & 3.60 & 2.58 & 3.91 & 3.03 & 4.23 & 3.36 & 4.55 & 3.81 & 4.84 & 4.21 \\
         DAN(s.64) & \textbf{3.14} & 48.91 & \textbf{3.46} & 57.81 & \textbf{3.75} & 67.69 & \textbf{4.10} & 77.08 & \textbf{4.42} & 87.03 & \textbf{4.75} & 97.26\\
         \bottomrule
    \end{tabular}
    \label{table2}
\end{table*}

\begin{table*}[t]\footnotesize
    \centering
    \caption{Results on TSPlib instances where longer computing time is allowed. }
    \vspace{-0.3cm}
    \begin{tabular}{l|p{0.9cm}p{0.9cm}|p{0.9cm}p{0.9cm}|p{0.9cm}p{0.9cm}|p{0.9cm}p{0.9cm}|p{0.9cm}p{0.9cm}}
    \toprule
         \multicolumn{1}{c|}{\multirow{2}{*}{Method}} & \multicolumn{2}{c|}{eil51} & \multicolumn{2}{c|}{eil76} & \multicolumn{2}{c|}{eil101} & \multicolumn{2}{c|}{kroa150} & \multicolumn{2}{c}{tsp225} \\
          & m=5 & m=10 & m=5 & m=10 & m=5 & m=10 & m=10 & m=20 & m=10 & m=20 \\
         \midrule
         LKH3 \hfill\cite{helsgaun2017extension} & 119 & 112 & 142 & 126 & 147 & 113 & 1554 & 1554 & 998 & 999 \\
         OR (600s) & 119 & 114 & 145 & 130 & 153 & 116 & 1580 & 1560  & 1068 & 1143 \\
         DAN (s.256) & 126 & 113 & 160 & 128 & 168 & 116 & 1610 & 1571 & 1111 & 1032 \\
         \bottomrule
    \end{tabular}
    \label{table3}
\end{table*}

We compare the performance of our model with both conventional and neural-based methods. For conventional methods, we test \textit{OR Tools}, evolutionary algorithm (EA), and self organizing maps (SOM)~\cite{lupoaie_som-guided_2019} on the same test set. \textit{OR Tools} initially gets a solution using meta-heuristic algorithms (path-cheapest-arc) and then further improves it by local search (2-opt)~\cite{gendreau_guided_2010} (denoted as OR).
We allow \textit{OR Tools} to run for a similar amount of time as our sampling variant, for fair comparison. Note that \textit{OR Tools} is always allowed to perform local search if there is computing time left after finding an initial solution. Exact algorithms need hours of time to solve one mTSP instances. Here, we report the results of Gurobi~\cite{Gurobi}, one of the state-of-the-art integer linear programming solver, from Hu et al.'s paper~\cite{hu2020reinforcement}, where 1 hour of computation was allowed for each instance. 
Table~\ref{table1} reports the average MinMax cost (lower is better) for small-scale mTSP instances (from 50 to 200 cities), as well as the average computing time per instance for each solver.

For neural-based methods, we report Park et al.'s results~\cite{park_schedulenet_2021} and Hu et al.'s results~\cite{hu2020reinforcement} from their papers, since they did not make their code available publicly.
Since Park et al.'s paper does not report the computing time of their approach, we leave the corresponding cells blank in Table~\ref{table1}.
Similarly, since Hu et al. did not provide any results for cases involving more than 100 cities or more than 10 agents, these cells are also left blank.
Note that the test sets used by Park et al. and Hu et al. are likely different from ours, since they have not been made public. 
However, the values reported here from their paper are also averaged over 500 instances under the same exact conditions as the ones used in this work. 
Finally, for completeness, we also test Kool et al.'s (TSP) model on our mTSP instances, by simply replacing our neural network structure with theirs in our distributed RL framework.

Table~\ref{table2} shows the average MinMax cost for large-scale mTSP instances (from 500 to 1000 cities), where the number of agents is fixed to $10$ (due to the limitation of Hu et al.'s model). When testing our sampling variant, we set $C = 100$ in the third decoder layer for efficient exploration (since the tour is much longer). Except DAN and Hu et al.'s model, no method can handle such large-scale mTSP within reasonable time, but we still report the results of \textit{OR Tools} from Hu et al.'s paper, as well as SOM results as the best-performing meta-heuristic algorithms for completeness.

Table~\ref{table3} shows the test results on TSPlib~\cite{tsplib} instances, a well-known benchmark library, where city distributions come from real-world data. There, we extend the computing time of \textit{OR Tools} to $600$s and increase the sample size of DAN to $256$, to yield solutions as optimal as possible. Note that the computing time of DAN never exceeds $100$s. LKH3 results are reported from~\cite{helsgaun2017extension}, where long enough computing time were allowed to yield exact solutions.


\subsection{Discussion}

We first notice that DAN significantly outperforms \textit{OR Tools} in larger-scale mTSP instances ($m>5$, $n\geq 50$), but is outperformed by \textit{OR Tools} in smaller-scale instances (as can be expected). In smaller-scale mTSP, \textit{OR Tools} can explore the search space sufficiently to produce near-optimal solutions. In this situation, DAN and all other decentralized methods considered find it difficult to achieve the same level of performance.

\begin{wrapfigure}{r}{0.55\textwidth}
    \vspace{-0.8cm}
    \centering
    \includegraphics[width=0.55\textwidth]{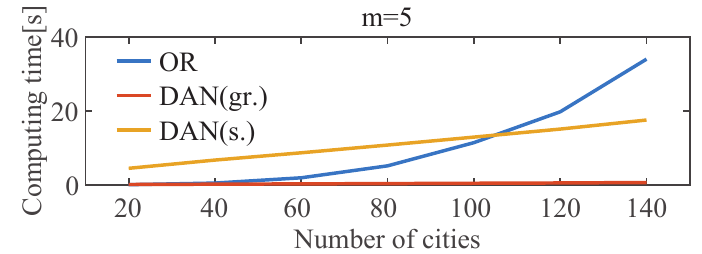}
    \vspace{-0.7cm}
    \caption{Planning time for the different solvers from $n=20$ to $n=140$ while $m=5$. The computing time of our model only increases linearly with respect to the number of cities, while the computing time of \textit{OR Tools} (without local search) increases exponentially.}
    \label{computing_time}
    \vspace{-0.7cm}
\end{wrapfigure}
However, DAN outperforms \textit{OR Tools} in larger-scale mTSP instances, where \textit{OR Tools} can only yield sub-optimal solutions within the same time budget.
In mTSP100($m\geq 10$), our sampling variant is 10\% better than \textit{OR Tools}. The advantage of our model becomes more significant as the scale of the instance grows, even when using our greedy variant. For instances on instances involving 500 or more cities, \textit{OR Tools} becomes impractical even when allowing up to 1800s per instance, while our greedy variant still outputs solutions with good quality in a matter of seconds.
In general, the computing time of our model increases linearly with the scale of the mTSP instance, as shown in Fig.~\ref{computing_time}.

Second, we notice that DAN's structure helps achieve better agent collaboration than the other decentralized dRL methods, thus yielding better overall results.
In particular, we note that simply applying a generic encoder-decoder structure (Kool et al.’s model) only provides mediocre to low-quality solutions across mTSP instances. This supports our claim that DAN's extended network structure is key in yielding drastically improved performance over this original work.
More generally, we note that all four dRL methods tested in our work (i.e., DAN, Hu et al.’s model, Park et al.’s model, and Kool et al.’s model) contain similar network structures as our city encoder and decoder.
Therefore, our overall better performance seem to indicate that our additional agent encoder and city-agent encoder are responsible for most of DAN’s increased performance.

Finally, we notice that DAN can provide near-optimal solutions for larger teams in medium-scale mTSP instances, while keeping computing times much lower than non-learning baselines (at least one order of magnitude). As shown in Table~\ref{table1} and ~\ref{table3}, although exact solvers cannot find reasonable mTSP solutions in seconds/minutes like \textit{OR Tools} and DAN, by running for long enough they still can guarantee near-optimal solutions. However,  DAN’s performance is still close (within 4\%) to LKH3 in problems involving 10 agents.


\section{Conclusion}
\label{DARS2022-DAN-conclusion}

This work introduced DAN, a decentralized attention-based neural network to solve the MinMax multiple travelling salesman (mTSP) problem. We approach mTSP as a cooperative task and formulate mTSP as a sequential decision making problem, where agents distributedly construct a collaborative mTSP solution iteratively and asynchronously. In doing so, our attention-based neural model allows agents to achieve implicit coordination to solve the mTSP instance together, in a fully decentralized manner. Through our results, we showed that our model exhibits excellent performance for small- to large-scale mTSP instances, which involve $50$ to $1000$ cities and $5$ to $20$ agents. Compared to state-of-the-art conventional baseline, our model achieves better performance both in terms of solution quality and computing time in large-scale mTSP instances, while achieving comparable performance in small-scale mTSP instances. We notice such a feature may make DAN more interesting for robotics tasks, where mTSP needs to be solved frequently and within seconds or a few minutes.

We believe that the developments made in the design of DAN can extend to more general robotic problems where agent allocation/distribution is key, such as multi-robot patrolling, distributed search/coverage, or collaborative manufacturing. We also acknowledge that robotic applications of mTSP may benefit from the consideration of real-life deployment constraints directly at the core of planning. We note that such constraints as agent capacity, time window, city demand were added to learning-based TSP solvers with minimal change in network structure~\cite{kool2018attention}. We believe that the same should hold true for DAN, and such developments will be the subject of future works as well.

\newpage
\bibliographystyle{spmpsci_unsrt}
\bibliography{ref2}

\end{document}